\documentclass{article}
\usepackage{spconf,amsmath,amssymb,graphicx}
\usepackage{multicol,multirow}
\usepackage{subcaption}
\usepackage{microtype}
\usepackage{arydshln}
\usepackage{comment}
\usepackage{xcolor}
\usepackage{booktabs}

\def\ie{\textit{i.e.}}
\def\eg{\textit{e.g.}}

\newcommand{\ec}[1]{\textcolor{black}{#1}}

\newcommand{\codebook}[1]{\scriptsize{\sc{\textcolor{teal}{#1}}}}

\title{FFT-based Selection and Optimization of Statistics for Robust Recognition of Severely Corrupted Images}
\name{Elena Camuffo$^{1,2,*}$, Umberto Michieli$^{1}$, Jijoong Moon$^{3}$, Daehyun Kim$^{3}$, Mete Ozay$^{1}$ \thanks{* Research completed during internship at Samsung Research UK.
}}

\address{$^{1}$ Samsung Research UK, $^{2}$ University of Padova, $^{3}$ Samsung Research Korea}
\begin{document}
%
\maketitle
\begin{abstract}
Improving model robustness in case of corrupted images is among the key challenges to enable robust vision systems on smart devices, such as robotic agents. Particularly, robust test-time performance is imperative for most of the applications.
This paper presents a novel approach to improve robustness of any classification model, especially on severely corrupted images. Our method (\textbf{FROST}) employs high-frequency features to detect input image corruption type, and select layer-wise feature normalization statistics. 
\textbf{FROST} provides the state-of-the-art results for different models and datasets, outperforming competitors on ImageNet-C by up to $37.1\%$ relative gain, improving baseline of $40.9\%$ mCE on severe corruptions.
\end{abstract}
\vspace{-0.5em}
\section{Introduction}
\label{sec:intro}
\vspace{-0.2em}

Achieving robustness of object recognition models on corrupted images is a challenging problem, which has been studied extensively in recent years \cite{hendrycks2019robustness,yucel2023hybridaugment,lim2023ttn,Chen_2021_ICCV,long2022frequency}. While the models have achieved impressive results on several benchmarks \cite{russakovsky2015imagenet, Krizhevsky2009LearningML}, recent works show that their performance is severely degraded when dealing with corrupted images. Addressing the robustness issues is important to ease the adoption of the models in practical applications on devices used in the wild, where such corruptions are frequently experienced.
We tackle this problem considering a collection of image distortions that are commonly observed in real-world natural images.
Object recognition using distorted image datasets can be posed as a dataset shift problem  \cite{Sun2017ImprovingRO}.
Although there has been substantial prior work investigating this problem, it is far from being fully understood, let alone solved. The most successful approaches make use of data augmentation \cite{wang2021augmax,autoaugment,Hendrycks2021PixMixDP,hendrycks2020augmix} and adversarial training \cite{bai2021transformers,long2022frequency,yucel2023hybridaugment}.
\begin{figure}[t]
    \centering
    \begin{minipage}{\linewidth}
    \centering
    \includegraphics[width=\textwidth]{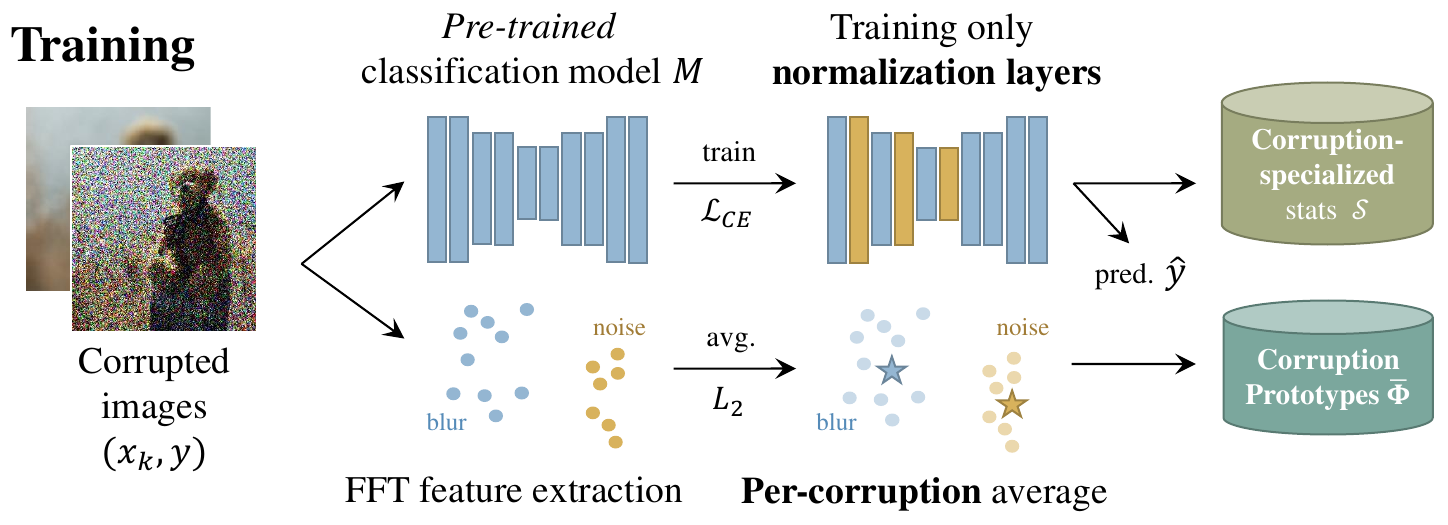}
    \end{minipage}\hfill
        \begin{minipage}{\linewidth}
    \centering
        \includegraphics[width=\textwidth]{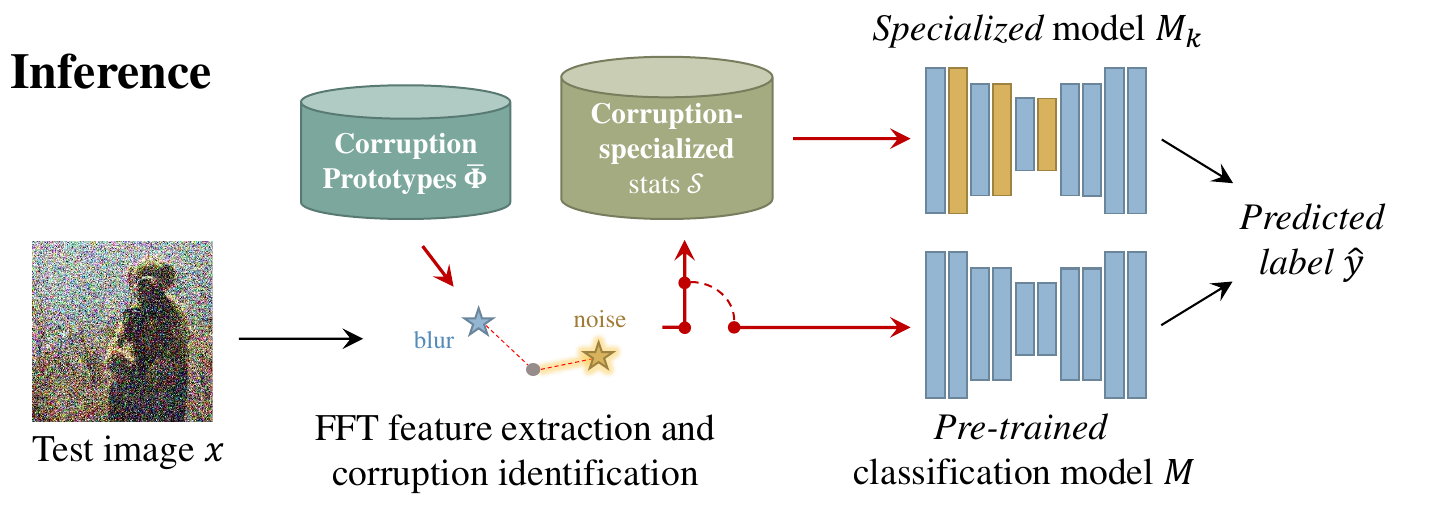}
    \end{minipage}
    \caption{\small Overall pipeline of our \textbf{FROST}. At \textit{training time}, we construct (i) corruption-specific prototypes using high-frequency FFT features and (ii) corruption-specific feature normalization statistics. At \textit{test time}, we extract FFT features and perform inference via prototype matching to select the most suitable statistics.}
    \label{fig:pipeline}
\end{figure}
Some methods focus on resolving shift of data statistics to improve model robustness, developing modules on top of feature normalization layers  \cite{lim2023ttn,Schneider2020ImprovingRA,Khurana2021SITASI}.
Recent studies leveraged frequency spectra insights to improve model robustness \cite{Chen_2021_ICCV,yucel2023hybridaugment,long2022frequency}. 
However, these methods often use cumbersome ensemble models \cite{Saikia_2021_ICCV} or formulate complex augmentation regimes \cite{Sun2022ASV,long2022frequency} at training time, with no possibility to adapt the models to test data. In addition, it is important to preserve, if not improve the accuracy of the models on clean samples while enhancing its robustness.

We improve robustness of models against corrupted images by selecting layer-wise feature normalization statistics depending on the corruption via high-frequency  features. 
At test-time, we determine the corruption present in an input image, by processing FFT features and matching them to corruption-specific prototypical features (prototypes). 
Then, we use codebooks mapping prototypes to corruption-specific statistics provided by normalization layers of models, to adapt the models to input image corruption. %
Our main contributions can be summarized as follows:
(1) we propose a novel test-time \underline{F}FT-based \underline{RO}bust \underline{ST}atistics selection method (\textbf{FROST}) based on a codebook mapping FFT features to corruption-specific statistics;
(2)	we achieve model robustness to image corruptions at test-time, making it useful for mobile and robotic applications in the wild;
(3)	we improve accuracy against competitors on ImageNet-C  \cite{hendrycks2019robustness}, especially in case of \textit{severe} corruptions, while preserving accuracy on clean data;
(4)	our solution is portable to different models and architectures, optimizing statistics of feature normalization layers (Batch/Layer Normalization), with very limited storage requirements. 

\section{Method}
\label{sec:method}
\vspace{-0.2em}
In this section, we introduce the details of our method. \textbf{FROST} (Fig.~\ref{fig:pipeline}) performs a 2-step approach: 
\ec{At \textit{training time}, FROST extracts high-frequency amplitudes from corrupted images, it aggregates them for images with the same corruption, and it builds a set of per-corruption feature prototypes. Then, it estimates corruption-specific (\textbf{Corr-S}) and corruption-generic (\textbf{Corr-G}) normalization layer parameters starting from a pretrained model.
At \textit{test time}, FROST identifies corruption types $\hat{k}$ present in the test images and uses a \textit{codebook} $\mathcal{C}$ to map such corruptions to normalization layers' parameters to minimize the recognition error. These normalization parameters $\mathcal{S}'_{\hat{k}}$ come from either corruption-generic or specific model, depending on the confidence of the model.}

\textbf{Background.}
Given a model $M$ that approximates ground truth labels $y \in \mathcal{Y}$ of samples  $\mathbf{x} \in \mathcal{X}$ using a training set $\mathcal{T} = \mathcal{X} \times \mathcal{Y}$. 
We define a corrupted image by ${\tilde{\mathbf{x}} = \mathbf{x} + \psi}$, where ${\mathbf{x} \in \mathbb{R}^{w \times h \times 3}}$ is a clean RGB image with the width $w$, height $h$ and corruption $\psi$. Our goal is to improve object recognition accuracy of $M$ on the corrupted images.

\begin{figure}[t]
    \centering 
    {\includegraphics[trim=0.4cm 0cm 0.4cm -0.6cm,clip,width=\linewidth]{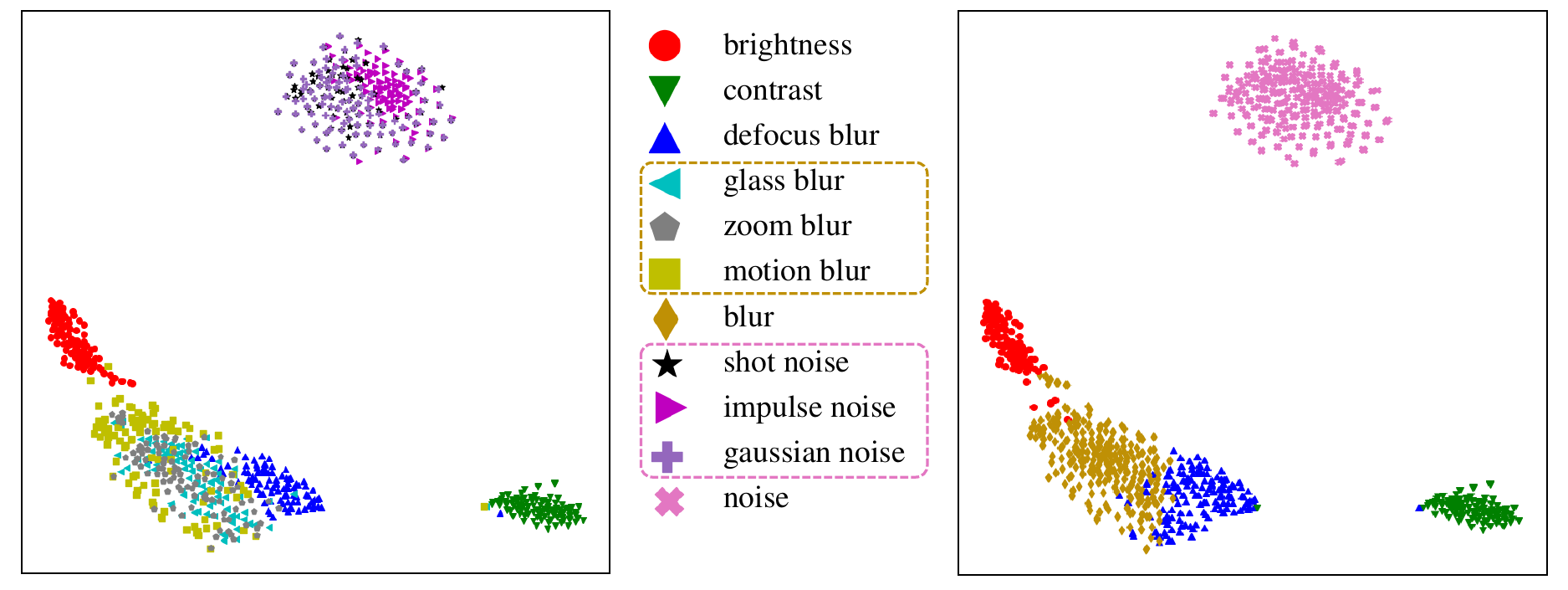}
    \subcaption*{$\ $ Original \hspace{4cm} k-means}}
    \caption{\small t-SNE projection of FFT features. Grouping indistinguishable clusters (\ie, \textit{noise} and \textit{blur} except \textit{defocus blur}),  no significant difference is visible between the ground truth and k-means ones. Our method produces FFT features similar to ground truth ones.}
    \label{fig:tsne}
\end{figure}

\textbf{Corruptions.} Previous works \cite{hendrycks2019robustness,wang2021augmax} showed that a real corruption $\psi$ can be approximated by a combination of synthetic corruptions. We use a subset
$\mathcal{K} = \{$\textit{Contrast}, \textit{Brightness}, \textit{Defocus Blur}, \textit{Glass Blur}, \textit{Motion Blur}, \textit{Zoom Blur}, \textit{Impulse Noise}, \textit{Shot Noise}, \textit{Gaussian Noise}$\}$ of the $9$ most common real corruptions defined in \cite{hendrycks2019robustness}.
We denote a synthetic corruption for the clean image $\mathbf{x}$ by $\psi_k^{\lambda}(\mathbf{x})$ such that $\mathbf{x} + \psi_k^{\lambda}(\mathbf{x}) \approx \tilde{\mathbf{x}}$ for $k \in \mathcal{K}$ (\eg, $k=$ \textit{Contrast}). The parameter $\lambda \in \{1,2,3,4,5\}$  is an integer number which defines corruption intensity depending on the degradation level, with $\lambda=1$ being the lowest and $\lambda=5$ the highest.

\begin{figure}[t]
    \centering
    {\includegraphics[trim=2cm 0.2cm 1.8cm 1.2cm,clip,width=\linewidth]{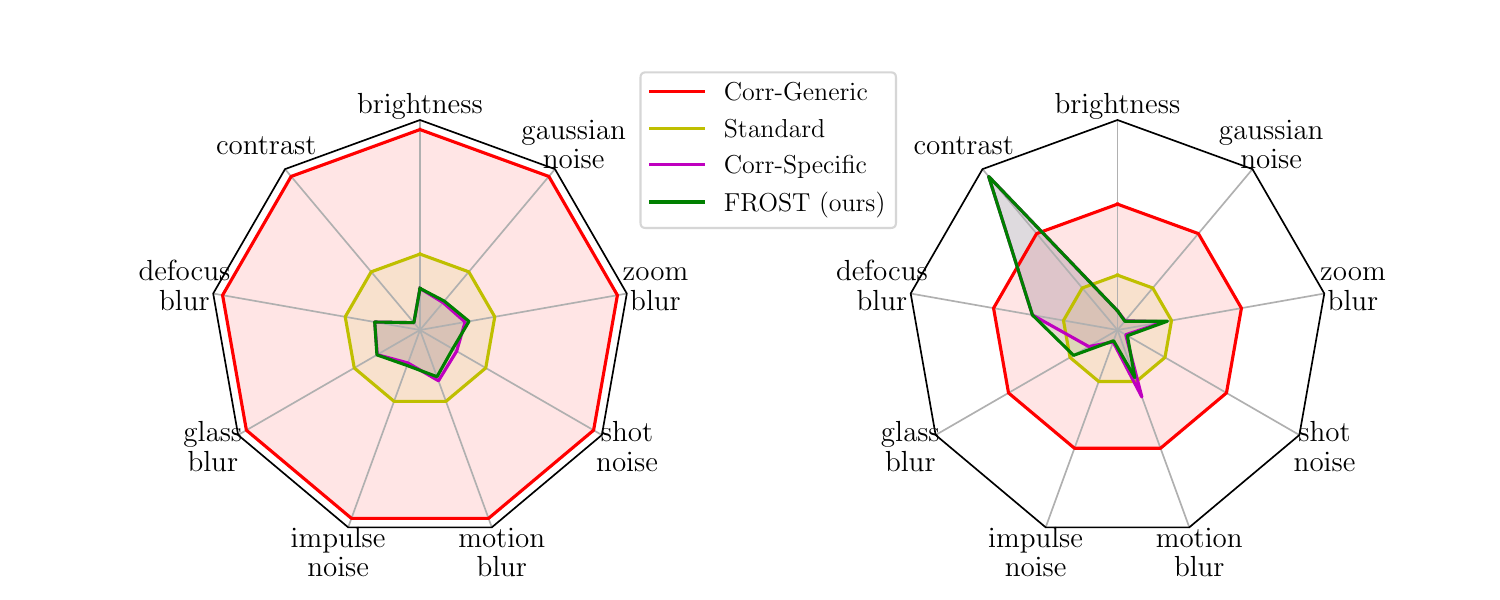}
        \subcaption*{$\ \ \ $  {Mean} \hspace{3.6cm} {Variance}}}
    \caption{\small Variation of normalization layers statistics between standard pretrained model (on clean data), corruption-generic $\mathcal{S}_{\mathcal{DA}}$ (\textbf{Corr-G}) model, corruption-specific (\textbf{Corr-S}) model, and normalization statistics of \textbf{FROST}, obtained aggregating them according to similar corruptions.} 
    \label{fig:bn}
\end{figure}

\begin{table*}[t]
    \centering \footnotesize
        \setlength{\tabcolsep}{2.2pt}
    \renewcommand{\arraystretch}{0.95}
    \begin{tabular}{lccccccccccccc}
    \toprule
Model & {FFT} & {BN} & \textbf{Contrast} & \textbf{Defocus B.} & \textbf{Glass B.$^{\dag}$} & \textbf{Motion B.$^{\dag}$} & \textbf{Zoom B.$^{\dag}$} & \textbf{Impulse N.$^{\ddag}$} & \textbf{Shot N.$^{\ddag}$} & \textbf{Gaussian N.$^{\ddag}$} & \textbf{Brightness} & \textbf{mCE} & \textbf{Error} \\ 
\midrule
        Standard & ~ & ~ & 79 & 82 & 90 & 84 & 80 & 82 & 80 & 79 & 65 & 80.1 & 23.9 \\ 
        Patch Uniform \cite{lopes2019improving} & ~ & ~ & 77 & 74 & 83 & 81 & 77 & 70 & 68 & 67 & 62 & 73.2 & 24.5 \\ 
        AA \cite{autoaugment} & ~ & ~ & 70 & 77 & 83 & 80 & 81 & 72 & 68 & 69 & 56 & 72.9 & 22.8 \\ 
        Random AA \cite{autoaugment} & ~ & ~ & 75 & 80 & 86 & 82 & 81 & 72 & 71 & 70 & 61 & 75.3 & 23.6 \\ 
        MaxBlur Pool \cite{pmlr-v97-zhang19a} & ~ & ~ & 68 & 74 & 86 & 78 & 77 & 76 & 74 & 73 & 56 & 73.6 & 23.0 \\ 
        SIN \cite{Rusak2020ASW} & ~ & ~ & 69 & 77 & 84 & 76 & 82 & 70 & 70 & 69 & 65 & 73.6 & 27.2 \\ 
        APR \cite{Chen_2021_ICCV} & $\checkmark$ & ~ & 61 & 72 & 86 & 72 & 79 & 64 & 68 & 62 & 58 & 69.1 & 25.5 \\ 
        BN \cite{Schneider2020ImprovingRA} & ~ & $\checkmark$ & 63 & 66 & 62 & 88 & 63 & 72 & 109 & 51 & 36 & 67.9 & | \\ 
        AugBN \cite{Khurana2021SITASI} & ~ & $\checkmark$ & 60 & 65 & 60 & 86 & 63 & 70 & 106 & 50 & 36 & 66.4 & | \\
        AugMix \cite{hendrycks2020augmix} & ~ & ~ & 58 & 70 & 80 & 66 & 66 & 67 & 66 & 65 & 58 & 66.2 & 22.4 \\ 
        PixMix \cite{Hendrycks2021PixMixDP} & ~ & ~ & 53 & 73 & 88 & 77 & 76 & 51 & 52 & 53 & 56 & 64.3 & 22.6 \\ 
        DA \cite{Hendrycks_2021_ICCV} & ~ & ~ & 64 & 63 & 75 & 69 & 75 & 45 & 47 & \textbf{46} & 55 & 59.9 & 23.4 \\
        HA \cite{yucel2023hybridaugment} & $\checkmark$ & ~ & 68 &  62 & 75 & 69 & 73 & 55 & 58 & 57 & 61 & 64.2 & 24.4 \\ 
        AugMix+DA+HA & $\checkmark$ & ~ & 54 & 52 & 66 & 54 & 65 & \textbf{44} & 46 & \textbf{46} & 53 & 53.3 & 24.9 \\
        \midrule
            \textbf{FROST}$_{0.1}$ (Ours) & $\checkmark$ & $\checkmark$ & \textbf{31} & \textbf{35} & 62 & 43 & 54 & 58 & 68 & 60 & \textbf{35} & 49.4 & 28.0 \\  
         \textbf{FROST}$_{0.2}$ (Ours) & $\checkmark$ & $\checkmark$ &  \textbf{31} & 40 & \textbf{57} & \textbf{36} & \textbf{48} & \textbf{44} & \textbf{45} & \textbf{46} & 41 & \textbf{43.0}  & 26.1 \\ 
         &  & & \codebook{44.8} & \codebook{54.5} & \codebook{74.7} & \codebook{74.2} & \codebook{77.9} & \codebook{41.0} & \codebook{38.1} & \codebook{44.3} & \codebook{84.4} & \codebook{59.3} \\  
    \bottomrule
    \end{tabular}
\caption{\small Analyses of the Corruption Error (CE) $\downarrow$ for the ResNet50 on the ImageNet-C, which is computed by an average of the single errors on the 5 corruptions and compared to the classification error obtained using AlexNet, as in \cite{hendrycks2019robustness}. \textbf{Error} $\downarrow$ is the error of the model on clean images. \textbf{mCE} $\downarrow$ is the classification error averaged on all the aforementioned corruptions. \textcolor{teal}{Codebook results} are reported in terms of accuracy $\uparrow$.}
\label{tab:res-in-t} \vspace{-0.5em}
\end{table*}

\begin{table*}[t]
    \centering \footnotesize
        \setlength{\tabcolsep}{2.2pt}
    \renewcommand{\arraystretch}{0.95}
    \begin{tabular}{lccccccccccccc}
    \toprule
Model & {FFT} & {BN} & \textbf{Contrast} & \textbf{Defocus B.} & \textbf{Glass B.$^{\dag}$} & \textbf{Motion B.$^{\dag}$} & \textbf{Zoom B.$^{\dag}$} & \textbf{Impulse N.$^{\ddag}$} & \textbf{Shot N.$^{\ddag}$} & \textbf{Gaussian N.$^{\ddag}$} & \textbf{Brightness} & \textbf{mCE} & \textbf{Error} \\ 
\midrule
        Standard & ~ & ~ & 73 & 95 & 110 & 102 & 85 & 94 & 96 & 95 & 58 & 89.7 & 23.9 \\
        BN \cite{Schneider2020ImprovingRA} & ~ & $\checkmark$ & 100 & 88 & 92 & 110 & 88 & 87 & 150 & 65 & 46 & 91.6 & |\\
        AugMix+DA+HA & $\checkmark$ & ~ & 78 & 68 & 85 & 73 & 79 & 54 & 64 & 58 & 56 & 68.4 & 24.9 \\
        AugBN \cite{Khurana2021SITASI} & ~ & $\checkmark$ & 60 & 65 & \textbf{60} & 86 & 63 & 70 & 106 & 50 & 36 & 66.4 & | \\
        \midrule
        \textbf{FROST}$_{0.2}$ (Ours) & $\checkmark$ & $\checkmark$ & \textbf{30} & 51 & 75 & \textbf{51} & \textbf{60} & 54 & \textbf{59} & 58 & \textbf{51} & 54.1  & 26.1  \\
        \textbf{FROST}$_{0.1}$ (Ours) & $\checkmark$ & $\checkmark$ & \textbf{30} & \textbf{44} & 75 & \textbf{51} & 61 & \textbf{40} & 60 & \textbf{45} & \textbf{33} & \textbf{48.8} & 28.0 \\ 
        & & & \codebook{98.0} & \codebook{92.4} & \codebook{90.8} & \codebook{85.5} & \codebook{86.8} & \codebook{83.0} & \codebook{46.2} & \codebook{79.0} & \codebook{80.9} & \codebook{87.5} \\
    \bottomrule
    \end{tabular}
\caption{\small Analyses of the Corruption Error (CE) $\downarrow$ for the ResNet50 on the ImageNet-C on \textit{severe} corruptions (\ie, severity levels $\lambda=4,5$), which is computed by an average of the single errors on the 2 corruptions and compared to the classification error obtained using AlexNet, as in \cite{hendrycks2019robustness}. \textbf{Error} $\downarrow$ is the error of the model on clean images. \textbf{mCE} $\downarrow$ is the mean corruption error.} 
\label{tab:res-in-s} \vspace{-1.5em}
\end{table*}

\medskip
\noindent\textit{Training. $\ $}
\textbf{FFT features extraction.} For each synthetic corruption in $\mathcal{K}$, we construct a set ${\mathcal{X}}_k^5$, by applying a corruption $\psi_k^5(\mathbf{x})$ to all the images $\mathbf{x} \in \mathcal{X}$. In this case, we use only the strongest corruption ($\lambda=5$) to obtain a better separation between features for different corruptions.
For each image $\tilde{\mathbf{x}}_k = \mathbf{x} + \psi_k^\lambda(\mathbf{x})$ with corruption $k$, we extract an FFT feature $\Phi_k=\mathfrak{F}_n(\tilde{\mathbf{x}}_k)$ by performing FFT $\mathfrak{F}(\cdot)$ \cite{cooley1965} on the input image, applying windowing operation to retain the first $n$ high-frequency components of the amplitude spectrum and flattening. In particular, $n$ is selected empirically as $n=15$, computing $\mathfrak{F}(\cdot)$ on images resized to $64\times64$.
Then, we average each set of features specific to corruption $k$ to obtain a prototype $\bar{\Phi}_{k} = \frac{1}{N} \sum_{\tilde{\mathbf{x}}_{k} \in \tilde{\mathcal{X}}_{k}}{\Phi_k}, \forall k\in \mathcal{K}$, with $N$ being the size of the training set. This can be done via a running average during training with no need to store all features into memory. We call this set $\bar{\Pi}$.
Analyzing the set of features $\{\Phi_k\}_{k \in \mathcal{K}}$ for different corruptions, we note that some results are very well clustered (\eg, \textit{Contrast}, \textit{Brightness} and \textit{Defocus Blur}), while others (\eg, \textit{Blur} types and \textit{Noise} distortions) are hardly separable.
To obtain a better clustering score, we compute k-means on the set $\{\Phi_k\}_{k \in \mathcal{K}}$ of FFT features (see Fig.~\ref{fig:tsne}).
This set is originally labeled with corruption-relative labels $L$ from $\mathcal{K}$. Let us define a new labeling $L^{*}$ obtained through k-means.
Setting empirically the number of clusters for the k-means to $5$, we obtain similar clustering score as for original labels. In particular, if we group together \textit{Blur} corruptions, and \textit{Noise} corruptions, we obtain a new labeling $L'$. Comparing $L^{*}$ with $L'$ via quantitative analysis, we get an adjusted random score \cite{Hubert1985ComparingP} of $89.1\%$ (meaning that the clusters are very similar). For this reason, we aggregate prototypes for features belonging to similar corruptions, obtaining a new set of corruptions ${\mathcal{K}'= \{\textit{Contrast}, \textit{Brightness}, \textit{Defocus}, \textit{Blur}, \textit{Noise}\}}$ with $5$ \textit{macro} corruptions. 
Also, we obtain a new set of \textit{macro} prototypes $\bar{\Pi}'$ averaging prototypes with really close corruptions.

\ec{Note that in real-world cases, high-frequency visual content could interfere with the corruption-related frequency content. In those cases, the algorithm can be extended adopting a multi-feature \cite{multifeat} or a multi-scale \cite{Li2020OnPO} FFT approach.}

\textbf{Estimation of corruption-specific statistics.}
We define by $S$ the set of statistics 
estimated at all normalization layers (Batch/Layer Normalization) in the recognition model $M$. These layers are \textit{storage-friendly} as they have only two 
parameters (scale $\gamma$ and shift $\beta$) that have shown to adapt differently to input images affected by different corruptions \cite{lim2023ttn,wang2021augmax}. Therefore, our purpose is to use it to improve recognition accuracy for corrupted images. 
First, we train a model $M$ (updating only normalization layers with $S$ parameters) on $\mathcal{T}$ performing data augmentation ($\mathcal{DA}$) on clean samples by adding $\psi_k^{\lambda}(\cdot)$. Image augmentations are selected according to a uniform distribution using original corruption functions for augmentations ($K=9$ in total) with severe corruptions only, \ie, $\lambda\sim \{4,5\}$. With this training, we obtain corruption-generic normalization statistics $S_{\mathcal{DA}}$.
Then, we train $M$ (updating only normalization layers with $S$ parameters) on $\mathcal{T}_k^{4,5}$ (\ie, $\mathcal{T}$ corrupted with corruption $k$, only using $\lambda\sim \{4,5\}$), producing $K$ different corruption-specific sets of normalization statistics $S_k$. 
According to \textit{macro} corruption grouping $\mathcal{K}'$, we average normalization statistics for indistinguishable corruptions obtaining $S'_k$ sets, one for each \textit{macro} corruption.

\begin{table*}[ht]
    \centering \footnotesize
    \setlength{\tabcolsep}{2.4pt}
    \renewcommand{\arraystretch}{0.95}
    \begin{tabular}{lcccccccccccccccccccccc}
    \toprule
    & \multicolumn{2}{c}{\textbf{Contrast}} & \multicolumn{2}{c}{\textbf{Defocus B.}} & \multicolumn{2}{c}{\textbf{Glass B.$^{\dag}$}} & \multicolumn{2}{c}{\textbf{Motion B.$^{\dag}$}} & \multicolumn{2}{c}{\textbf{Zoom B.$^{\dag}$}} & \multicolumn{2}{c}{\textbf{Impulse N.$^{\ddag}$}} & \multicolumn{2}{c}{\textbf{Shot N.$^{\ddag}$}} & \multicolumn{2}{c}{\textbf{Gaussian N.$^{\ddag}$}} & \multicolumn{2}{c}{\textbf{Brightness}} & \multicolumn{3}{c}{\textbf{Average}} & \textbf{Clean} \\
     & $\lambda_4$ & $\lambda_5$ & $\lambda_4$ & $\lambda_5$ & $\lambda_4$ & $\lambda_5$ & $\lambda_4$ & $\lambda_5$ & $\lambda_4$ & $\lambda_5$ & $\lambda_4$ & $\lambda_5$ & $\lambda_4$ & $\lambda_5$ & $\lambda_4$ & $\lambda_5$ & $\lambda_4$ & $\lambda_5$ & $\lambda_4$ & $\lambda_5$ &  \textbf{Tot}\\
    \midrule
        \textbf{Corr-S} & 69.9 & 67.4 & 65.7 & 63.3 & 65.9 & 60.4 & 68.1 & 66.9 & 64.3 & 62.0 & 67.3 & 62.7 & 66.5 & 63.6 & 66.5 & 61.7 & 69.0 & 66.9 & 67.0 & 63.9 & 65.1 & 73.1 \\
        \midrule
    B & 13.0  & 4.1 & 25.8 & 17.0 & 32.3 & 19.2 & 23.8 & 17.6 & 24.1 & 18.3 & 4.8 & 1.6 & 4.6 & 1.8 & 5.0 & 1.6 & 44.3 & 32.6 & 19.7 & 12.6 & 13.4 & \textbf{73.1} \\
    B + $\mathcal{DA}$ & 60.2 & 52.6 & 59.1 & 54.6 & \underline{62.1} & \underline{55.7} & \underline{62.0} & \underline{57.4} & \textbf{60.2} & \textbf{56.2} & \underline{64.2} & 58.9 & \textbf{63.5} & 60.0 & \underline{63.7} & 57.9 & 57.0 & 52.8 & \underline{61.3} & \underline{56.2} & \underline{59.3} & \underline{72.2} \\
B + $\mathcal{C}$ & \underline{62.2} & \underline{59.4} & \underline{62.4} & \underline{60.4} & 53.7 & 48.4 & 52.1 & 41.6 & 49.1 & 43.8 & 54.1 & \underline{61.6} & 60.6 & \underline{62.1} & 51.0 & \underline{60.7} & \underline{65.0} & \underline{60.3} & 56.7 & 55.4 & 56.0 & 70.0 \\ 
    & \codebook{82.7} & \codebook{81.2} & \codebook{92.6} & \codebook{100} & \codebook{89.7} & \codebook{94.4} & \codebook{88.2} & \codebook{92.5} & \codebook{77.9} & \codebook{99.8} & \codebook{84.6} & \codebook{77.1} & \codebook{90.9} & \codebook{98.6} & \codebook{84.2} & \codebook{86.7} & \codebook{72.0} & \codebook{99.4} & \codebook{84.8} & \codebook{92.0} & \codebook{88.4} \\
    B + $\mathcal{DA}$ + $\mathcal{C}$ & \textbf{68.7} & \textbf{66.6} & \textbf{65.1} & \textbf{62.7} & \textbf{62.6} & \textbf{55.8} & \textbf{63.6} & \textbf{57.5} & \underline{60.0} & \underline{56.1} & \textbf{67.2} & \textbf{62.0} & \underline{61.5} & \textbf{62.3} & \textbf{66.0} & \textbf{62.0} & \textbf{66.7} & \textbf{63.8} & \textbf{64.6} & \textbf{61.0} & \textbf{62.5} & 69.1 \\
     & \codebook{92.6} & \codebook{100} & \codebook{89.7} & \codebook{94.4} & \codebook{88.2} & \codebook{92.5} & \codebook{84.6} & \codebook{77.1} & \codebook{84.2} & \codebook{86.7} & \codebook{77.9} & \codebook{98.7} & \codebook{90.9} & \codebook{97.7} & \codebook{72.0} & \codebook{99.4} & \codebook{82.7} & \codebook{81.2} & \codebook{84.8} & \codebook{92.0} & \codebook{89.2} \\
    \bottomrule
    \end{tabular}
    \caption{\small Accuracy $\uparrow$ using the ResNet18 with the Tiny-ImageNet. We report the results for a standard model pretrained on clean images (B), \textbf{Corr-G} model (B + $\mathcal{DA}$), Codebook applied only on top of the standard model (B + $\mathcal{C}$) and \textbf{FROST} using the standard model (B + $\mathcal{DA}$ + $\mathcal{C}$). \textcolor{teal}{Codebook results} are reported below the relative experiment. \textit{Blur} (\dag) corruptions and \textit{Noise} (\ddag) corruptions share the same model, respectively. \textbf{Bold} indicates the best result and  \underline{underline} indicates the second best result. $\lambda=4$, $\lambda=5$ are shorten with $\lambda_4,\lambda_5$. \textbf{Tot} is the total average on all severe corruptions.} 
    \label{tab:results-percorr} \vspace{-1.5em}
\end{table*}

\begin{table}[ht]
    \centering \footnotesize
    \setlength{\tabcolsep}{2.3pt}
    \renewcommand{\arraystretch}{0.95}
    \begin{tabular}{lcccccc}
    \toprule
    \textbf{Model} & \textbf{Dataset} & \textbf{Stats} &  \textbf{Corr-G} & \textbf{FROST} & \textbf{Clean} \\
    \midrule
    ResNet18 \cite{resnet} & Tiny-ImageNet \cite{Krizhevsky2009LearningML} & BN & 59.3 & 62.5 \ (\codebook{89.2}) & 69.1\\
    ResNet50 \cite{resnet} & Tiny-ImageNet \cite{Krizhevsky2009LearningML} & BN
    & 66.7 & 68.6 \ (\codebook{88.6}) & 74.0\\
    ResNet101 \cite{resnet} & Tiny-ImageNet \cite{Krizhevsky2009LearningML} & BN & 
    74.9 & 76.5 \ (\codebook{84.9}) & 81.5\\
    {ViT-B} \cite{dosovitskiy2020vit} & Tiny-ImageNet \cite{Krizhevsky2009LearningML} & LN & 
    71.6 & 73.7 \ (\codebook{87.2}) & 79.0 \\
    {ViT-L} \cite{dosovitskiy2020vit} & Tiny-ImageNet \cite{Krizhevsky2009LearningML} & LN & 
    82.5 & 84.6 \ (\codebook{85.9}) & 85.2 \\
    \midrule
    ResNet18 \cite{resnet} & CIFAR10 \cite{Krizhevsky2009LearningML} & BN & 
      53.5 & 54.2 \ (\codebook{86.8}) & 70.1 \\ 
    ResNet18 \cite{resnet} & CIFAR100 \cite{Krizhevsky2009LearningML} & BN & 
     31.6 & 32.1 \ (\codebook{83.6}) & 46.6\\
    \bottomrule
    \end{tabular}
    \caption{\small Average accuracy $\uparrow$ on the corruption generic (\textbf{Corr-G}) model, \textbf{FROST} on corruptions and \textbf{Clean} data, using different datasets and networks. \textcolor{teal}{Codebook results} are reported in brackets.}
    \label{tab:results}
\end{table}

\medskip
\noindent\textit{Inference. $\ $}
At test time, we use prototypical features $\bar{\Pi}'$ as \textit{keys} of a codebook $\mathcal{C}$ to select the best set $S^{*}$.

\textbf{Prototype matching.} We perform inference on each test image $\tilde{\mathbf{x}}_u$ with unknown corruption $u$. 
First, we extract feature $\Phi_u = \mathfrak{F}_n(\tilde{\mathbf{x}}_u)$, retaining the first $n$ high-frequency components of the FFT amplitude spectrum.
Then, we compute the probability that image is corrupted with corruption $k$ such that $p(u=k) = \text{dist}_{\ell_2}(\Phi_u, \bar{\Phi}_k)$ for each corruption $k\in \mathcal{K}'$ using $\ell_2$ distance.
Note that a test image can also be non-corrupted; we will explain how this case is handled in the next paragraph.

\textbf{Selection of statistics.}
We use probability scores in order to select the most suitable set of normalization statistics $S^{*}$ via our codebook $\mathcal{C}$, and apply it on top of the model $M$ to enhance object recognition capabilities.
First, we determine whether the corruption is uncertain, by applying a thresholding operation on the first two most likely corruptions. 
We define $\hat{k}_1$ and $\hat{k}_2$ as the most likely and second most likely estimated corruptions. If $| p(u=\hat{k}_1) – p(u=\hat{k}_2) | < T$, then we use corruption-generic normalization statistics $\mathcal{S}_{\mathcal{DA}}$. Otherwise, we use corruption-specific normalization statistics $S'_{\hat{k}_1}$. 
In this case, $T$ is selected empirically (comparing distance values).
Note that clean images are generally mapped to $\mathcal{S}_{\mathcal{DA}}$; however they have intrinsic noise and sometimes using $S'_{\hat{k}_1}$ can be beneficial.
We remark that corruptions share the same normalization parameters in the standard pretrained model and in the \textbf{Corr-G} model $\mathcal{S}_{\mathcal{DA}}$. Instead, each corruption has its own set of normalization layer parameters in the \textbf{Corr-S} model, and aggregation of the \textbf{FROST} \textit{macro} corruptions provides a good approximation of it which is more convenient for corruption identification via FFT (see Fig.~\ref{fig:bn}). 

\vspace{-0.5em}
\section{Experimental Analyses}
\label{sec:results}
\vspace{-0.2em}

\textbf{ImageNet-C.} 
We evaluate our approach on the ImageNet-C dataset as done by compelling works. For this evaluation, \textbf{FROST} has been trained on the ImageNet, using a model with suitable data augmentation. In particular, we apply a codebook on top of a baseline model which is trained with AugMix  \cite{hendrycks2020augmix} + DA \cite{Hendrycks_2021_ICCV} + HA \cite{yucel2023hybridaugment}. 
We tested \textbf{FROST} with no further training for $S_{\mathcal{DA}}$, (i) $T=0.1$, and (ii) $T=0.2$. 
In Tabs.~\ref{tab:res-in-t},\ref{tab:res-in-s}, we report results for the Corruption Error (CE) (as defined in \cite{hendrycks2019robustness}), on all the severity levels and on severe corruptions only, respectively.
We report the \textit{codebook accuracy} for each corruption, defined as the percentage of correct codebook mappings.
\textbf{FROST} is compared with the other state-of-the-art approaches; particularly, with approaches using (i) Fourier Transform and (ii) normalization statistics. These approaches are marked in the first two columns of the tables. 
We show that our approach outperforms the other methods, gaining up to $37.1\%$ mCE on all corruptions and $40.9\%$ mCE on severe only. Specifically, if we compare \textbf{FROST} with the model used as baseline for augmentations (AugMix + DA + HA), it is shown that we can boost the mCE up to $10.3\%$ and $19.6\%$ for all and severe; moreover, results are dependent on $T$: a more intensive utilization of the codebook ($T=0.1$) is to be preferred in case of strong corruptions.
Indeed, for all corruptions, we have an overall improvement, which increases with the higher threshold $T=0.2$; for severe only, instead, the improvement is clear with any threshold (\ie, the network is more confident about which corruption is affecting the image).
This behavior is reflected by the codebook accuracy which is higher for stronger corruptions. 

\textbf{Ablation Studies.} We report in Tab.~\ref{tab:results-percorr} an ablation study on \textbf{FROST} components, to show their contribution to the accuracy. We trained a ResNet18 on the Tiny-ImageNet, and evaluated it on the validation set which are both corrupted with severe corruptions ($\lambda=4,5$). \textbf{Corr-S} shows good results for all the corruptions, but it is not feasible in real-world situations where we do not know a priori which corruption affects the test image.
The model pretrained on clean images (B) shows poor results on all components, but using \textbf{Corr-G} or codebook $\mathcal{C}$ boost results.
We show that our codebook $\mathcal{C}$ performs better than \textbf{Corr-G} in most of the cases. However, for less corrupted images ($\lambda=4$), we have few cases where \textbf{Corr-G} obtains higher accuracy. 
For this reason, we set the uncertainty threshold $T$, and let the codebook decide whether to use the generic or the specific model (as proved by results in Tabs.~\ref{tab:res-in-t},\ref{tab:res-in-s}). 
In this case, we set $T=0.1$, limiting the utilization of $S_{\mathcal{DA}}$; this results in optimal performance on the best clustered corruptions (\ie, \textit{Contrast}, \textit{Brightness}), but improvements are achieved also for other corruptions. A codebook performs well, when the corresponding corruption is clearly identifiable, thus it maximizes utilization when dealing with strong corruptions.

Finally, we show that \textbf{FROST} is applicable to any dataset or architecture. 
 Tab.~\ref{tab:results} shows the results in comparison with 5 different architectures, 3 different datasets, layer normalization (LN) and batch normalization (BN). The standard validation set with corruptions is used for evaluation. The results show that \textbf{FROST} is generalizable to all of them; specifically, attention must be given to ViT-B and ViT-L, where we observe that it is suitable also for LN statistics.
 As for memory occupation, the capacity required for storing statistics of the normalization layers for all 5 corruptions plus the storage required for the 5 prototypes results in an average increment of just 0.7\% on the total architecture size.
 
\vspace{-0.5em}
\section{Conclusion}
\label{sec:conclusion}
\vspace{-0.2em}

In this paper we proposed a novel test-time optimization approach for robust classification of severely corrupted images. High-frequency magnitude spectrum is exploited to select the most likely corruption and select layer-wise normalization statistics according to that. Experimental results show that our approach improves mCE on the ImageNet-C outperforming many SOTA approaches. Further, it shows suitability for different models and datasets, preserving clean accuracy.

\vfill

\bibliographystyle{IEEEbib}
\bibliography{refs}

\begin{thebibliography}{10}

\bibitem{hendrycks2019robustness}
Dan Hendrycks and Thomas Dietterich,
\newblock ``Benchmarking neural network robustness to common corruptions and
  perturbations,''
\newblock {\em ICLR}, 2019.

\bibitem{yucel2023hybridaugment}
Mehmet~K. Yucel, Ramazan~G. Cinbis, and Pinar Duygulu,
\newblock ``Hybridaugment++: Unified frequency spectra perturbations for model
  robustness,''
\newblock in {\em ICCV}, 2023.

\bibitem{lim2023ttn}
Hyesu Lim, Byeonggeun Kim, Jaegul Choo, and Sungha Choi,
\newblock ``{TTN: A Domain-Shift Aware Batch Normalization in Test-Time
  Adaptation},''
\newblock in {\em ICLR}, 2023.

\bibitem{Chen_2021_ICCV}
Guangyao Chen, Peixi Peng, Li~Ma, Jia Li, Lin Du, and Yonghong Tian,
\newblock ``Amplitude-phase recombination: Rethinking robustness of
  convolutional neural networks in frequency domain,''
\newblock in {\em ICCV}, 2021.

\bibitem{long2022frequency}
Yuyang Long, Qilong Zhang, Boheng Zeng, Lianli Gao, Xianglong Liu, Jian Zhang,
  and Jingkuan Song,
\newblock ``Frequency domain model augmentation for adversarial attack,''
\newblock in {\em ECCV}, 2022.

\bibitem{russakovsky2015imagenet}
Olga Russakovsky, Jia Deng, Hao Su, Jonathan Krause, et~al.,
\newblock ``Imagenet large scale visual recognition challenge,''
\newblock {\em IJCV}, 2015.

\bibitem{Krizhevsky2009LearningML}
Alex Krizhevsky, Geoffrey Hinton, et~al.,
\newblock ``Learning multiple layers of features from tiny images,''
\newblock 2009.

\bibitem{Sun2017ImprovingRO}
Zhun Sun, Mete Ozay, and Takayuki Okatani,
\newblock ``Improving robustness of feature representations to image
  deformations using powered convolution in cnns,''
\newblock in {\em CVPR}, 2017.

\bibitem{wang2021augmax}
Haotao Wang, Chaowei Xiao, Jean Kossaifi, Zhiding Yu, Anima Anandkumar, and
  Zhangyang Wang,
\newblock ``Augmax: Adversarial composition of random augmentations for robust
  training,''
\newblock in {\em NeurIPS}, 2021.

\bibitem{autoaugment}
Ekin~D. Cubuk, Barret Zoph, Dandelion Mané, Vijay Vasudevan, and Quoc~V. Le,
\newblock ``Autoaugment: Learning augmentation strategies from data,''
\newblock in {\em CVPR}, 2019.

\bibitem{Hendrycks2021PixMixDP}
Dan Hendrycks, Andy Zou, Mantas Mazeika, Leonard Tang, Bo~Li, Dawn~Xiaodong
  Song, and Jacob Steinhardt,
\newblock ``Pixmix: Dreamlike pictures comprehensively improve safety
  measures,''
\newblock {\em CVPR}, 2021.

\bibitem{hendrycks2020augmix}
Dan Hendrycks, Norman Mu, Ekin~D. Cubuk, Barret Zoph, Justin Gilmer, and Balaji
  Lakshminarayanan,
\newblock ``{AugMix}: A simple data processing method to improve robustness and
  uncertainty,''
\newblock {\em ICLR}, 2020.

\bibitem{bai2021transformers}
Yutong Bai, Jieru Mei, Alan Yuille, and Cihang Xie,
\newblock ``Are transformers more robust than cnns?,''
\newblock in {\em NeurIPS}, 2021.

\bibitem{Schneider2020ImprovingRA}
Steffen Schneider, Evgenia Rusak, Luisa Eck, Oliver Bringmann, Wieland Brendel,
  and Matthias Bethge,
\newblock ``Improving robustness against common corruptions by covariate shift
  adaptation,''
\newblock {\em NeurIPS}, 2020.

\bibitem{Khurana2021SITASI}
Ansh Khurana, S.~Paul, Piyush Rai, Soma Biswas, and Gaurav Aggarwal,
\newblock ``Sita: Single image test-time adaptation,''
\newblock {\em ArXiv}, 2021.

\bibitem{Saikia_2021_ICCV}
Tonmoy Saikia, Cordelia Schmid, and Thomas Brox,
\newblock ``Improving robustness against common corruptions with frequency
  biased models,''
\newblock in {\em ICCV}, 2021.

\bibitem{Sun2022ASV}
Jiachen Sun, Akshay Mehra, Bhavya Kailkhura, et~al.,
\newblock ``A spectral view of randomized smoothing under common corruptions:
  Benchmarking and improving certified robustness,''
\newblock in {\em ECCV}, 2022.

\bibitem{lopes2019improving}
Raphael~Gontijo Lopes, Dong Yin, Ben Poole, Justin Gilmer, and Ekin~D. Cubuk,
\newblock ``Improving robustness without sacrificing accuracy with patch
  gaussian augmentation,''
\newblock {\em ArXiv}, 2019.

\bibitem{pmlr-v97-zhang19a}
Richard Zhang,
\newblock ``Making convolutional networks shift-invariant again,''
\newblock in {\em ICML}, 2019.

\bibitem{Rusak2020ASW}
Evgenia Rusak, Lukas Schott, Roland~S. Zimmermann, et~al.,
\newblock ``A simple way to make neural networks robust against diverse image
  corruptions,''
\newblock in {\em ECCV}, 2020.

\bibitem{Hendrycks_2021_ICCV}
Dan Hendrycks, Steven Basart, Norman Mu, Saurav Kadavath, et~al.,
\newblock ``The many faces of robustness: A critical analysis of
  out-of-distribution generalization,''
\newblock in {\em ICCV}, 2021.

\bibitem{cooley1965}
James~W. Cooley and John~W. Tukey,
\newblock ``An algorithm for the machine calculation of complex fourier
  series,''
\newblock {\em MCOM}, 1965.

\bibitem{Hubert1985ComparingP}
Lawrence~J. Hubert and Phipps Arabie,
\newblock ``Comparing partitions,''
\newblock {\em Journal of Classification}, 1985.

\bibitem{multifeat}
Zheng Wang, Yanwei Zhao, and Jiacheng Chen,
\newblock ``Multi-scale fast fourier transform based attention network for
  remote-sensing image super-resolution,''
\newblock {\em IEEE Journal of Selected Topics in Applied Earth Observations
  and Remote Sensing}, 2023.

\bibitem{Li2020OnPO}
Bin Li, Zhikang Jiang, and Jie Chen,
\newblock ``On performance of multiscale sparse fast fourier transform
  algorithm,''
\newblock {\em Circuits, Systems and Signal Processing}, 2020.

\bibitem{resnet}
Kaiming He, Xiangyu Zhang, Shaoqing Ren, and Jian Sun,
\newblock ``Deep residual learning for image recognition,''
\newblock in {\em CVPR}, 2016.

\bibitem{dosovitskiy2020vit}
Alexey Dosovitskiy, Lucas Beyer, Alexander Kolesnikov, Dirk Weissenborn,
  et~al.,
\newblock ``An image is worth 16x16 words: Transformers for image recognition
  at scale,''
\newblock {\em ICLR}, 2021.

\end{thebibliography}

\end{document}